\documentclass[a4paper]{article}

\usepackage[english]{babel}
\usepackage[utf8x]{inputenc}
\usepackage[T1]{fontenc}

\usepackage[a4paper,top=3cm,bottom=2cm,left=3cm,right=3cm,marginparwidth=1.75cm]{geometry}

\usepackage{amsmath}
\usepackage{graphicx}
\usepackage[colorinlistoftodos]{todonotes}
\usepackage[colorlinks=true, allcolors=blue]{hyperref}

\usepackage{hyperref} 
\usepackage{authblk}

\title{LARNN: Linear Attention Recurrent Neural Network}
\author{Guillaume Chevalier}
\affil{Dept. of Computer Science and Software Engineering,\\
Laval University, Quebec, Canada\\}

\begin{document}
\maketitle

\begin{abstract}
The Linear Attention Recurrent Neural Network (LARNN) is a recurrent attention module derived from the LSTM cell and ideas from the consciousness Recurrent Neural Network (RNN). Yes, it LARNNs. The LARNN uses attention on its past cell state values for a limited window size $k$. The formulas are also derived from the Batch Normalized Long Short-Term Memory (BN-LSTM) cell and the Transformer Network for its Multi-Head Attention Mechanism. The Multi-Head Attention Mechanism is used inside the cell such that it can query its own $k$ past values with attention with restricted windowing on the $k$ most recent previous cell state. This has the effect of augmenting the rank of the tensor with the attention mechanism, such that the cell can perform complex queries to question its previous inner memories, which should augment the long short-term effect of the memory. With a clever trick, the LARNN cell with attention can be easily used inside a loop on the cell state, just like how any other Recurrent Neural Network (RNN) cell can be looped linearly through time series. This is due to the fact that its state, which is looped upon throughout time steps within time series, stores the inner states in a "first in, first out" queue which contains the $k$ most recent states and on which it is easily possible to add static positional encoding when the queue is represented as a tensor. This neural architecture yields better results than the plain, vanilla LSTM cells. It can obtain results of 91.924\% for the test accuracy, compared to the previously attained 91.653\% using vanilla LSTM cells. Note that this is not to compare to other research, where up to 93.349\% is obtained, but by costly using 18 LSTM cells rather than with 2 to 3 cells as analyzed here and in comparison. Finally, an interesting discovery is made, such that adding activation within the multi-head attention mechanism's linear layers can yield better results in the context researched hereto. 
\end{abstract}

\section{Introduction}

\subsection{Prior Art}

It have always been hard to replace the LSTM \cite{hochreiter1997long} since its discovery by Sepp Hochreiter and Jürgen Schmidhuber, that it'd be for language models or other kind of time series processing with neural networks, such as sensors signal. \\

Recently, Attention Mechanisms \cite{bahdanau2014neural} have been proven quite useful for Neural Machine Translation (NMT) when paired with RNNs. However, Attention Mechanisms are so good that they recently have been used alone without any kind of RNN nor Convolutional Neural Network (CNN). This marks the apparition of the Transformer Network \cite{vaswani2017attention}, claiming by the title of the paper that Attention Is All You Need. Since then, people started to believe that RNNs could be discarded in favor of attention mechanisms. \\

On the other side, there is Yoshua Bengio with his paper about The Consciousness Prior \cite{bengio2017consciousness}, in which he express consciousness as being recurrent through time, introducing the consciousness RNN which, by any means, may retain fragments of the input representations through time in the RNN cell as if it was attention. \\

\subsection{Proof that RNNs are here to stay}

However, RNNs can't be replaced: they are $O(n)$, while attention mechanisms are $O(n^2)$, where $n$ here represents the temporal axis along the processed time series. Those are different data structures, and it is not true that attention models can fully replace recurrent models. And even if the Attention Mechanisms would be stacked hierarchically (like rescursive pooling) with small windows - similarly to how a Wavenet \cite{van2016wavenet} would use convolutions with windowing) - at best this neural architecture would require $O(n log(n))$ time to process its input time series. The advantages of $O(n^2)$ attention only appears when the sequence length is small enough to be computed in $O(1)$ within a huge graphic card having enough memory and cores to process the whole thing all at once. All that being said, RNNs can't simply disappear: it's a fundamental linear data structure. Articles claiming the fall of RNN have no reason to exist and twists the facts. We shall not think of apples as oranges. 

\subsection{Logical reasoning leading to the idea of the LARNN}

\begin{figure}
\centering
\includegraphics[width=1\textwidth]{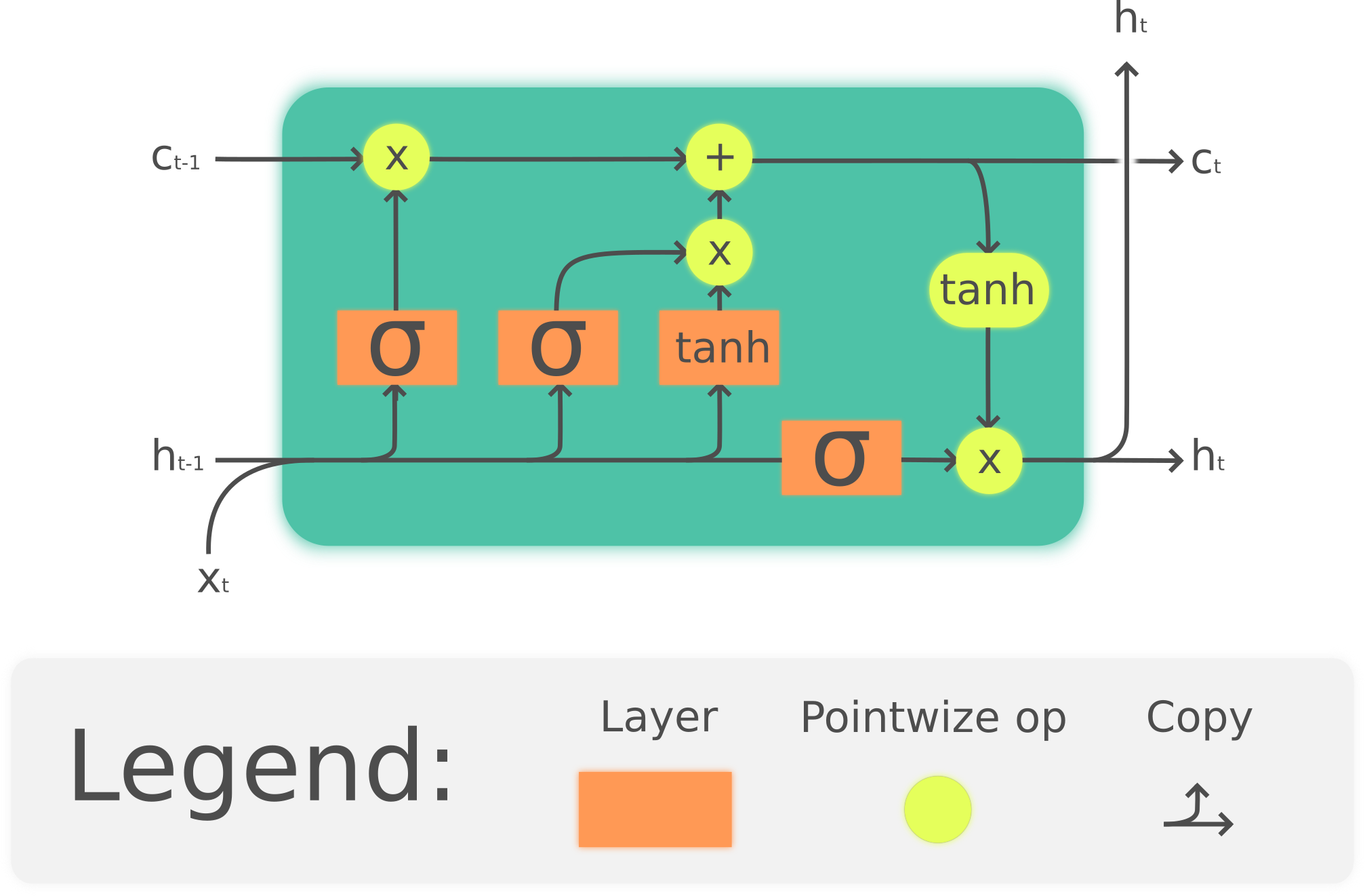}
\caption{\label{fig:lstm}The LSTM cell, which can be substituted by a consciousness RNN as introduced by Yoshua Bengio. This visualization is freely available and is licensed under the CC-BY License, by Guillaume Chevalier. For more information, visit \url{https://github.com/guillaume-chevalier/Linear-Attention-Recurrent-Neural-Network/tree/master/inkscape_drawings}.}
\end{figure}

Althought Self-Attention Mechanisms are there to stay, it is not the case that RNNs can be fully replaced. This is supported by Yoshua Bengio, as in his paper about The Consciousness Prior \cite{bengio2017consciousness}, where he formalizes that consciousness evolves linearly through time, just like an RNN or like an LSTM such as seen in Figure \ref{fig:lstm}. In his (yet still very underrated paper), he introduces the consciousness RNN, which is defined with the function $C$ as such:

$$c_t = C(h_t,~c_{t−1},~z_t)$$

where $z_t$ is a random source of noise, $c_t$ the conscious state, $c_{t-1}$ the previous conscious state, and $h_t$ a form of inputs' representation state. Overall, it it stated that $c_t$ is a form of attention which picked elements from $h_t$. That being said, it's quite reasonable to keep attentional neural architectures to be linear in time to keep the consciousness flowing through time, linearly. That is: using Attention Mechanisms within RNNs! \\

Why wouldn't such a consciousness be able to have attention over itself to digest information? For example, let's derive his equation such as to form a LARNN function with a consciousness $C$ which does not only examines its immediate past state: 

$$c_t = C(h_t,~c_{t−1},~c_{t−2},~c_{t−3},~...,~c_{t−k},~z_t)$$

where $k$ is an attention window which restricts how far the LARNN can see itself back in time, and on which how gradients can flow. This is now of complexity $O(n k^2)$ in time rather than $O(n)$, but the fact that $k$ is set to be a constant is pleasing. This constant is the clever trick described in the abstract. In practice, a queue can be used, and could have as first item a tensor filled with zeros, a random value, or an embedding encoding the initial setup. The queue can then be filled through time steps and is self-contained as a single tensor which changes with every new state, dropping too old states when the queue is full. This reduces the previous equation to: 

$$v_t = [[c_{t−1},~c_{t−2},~c_{t−3},~...,~c_{t−k}]]$$
$$c_t = C(h_t,~v_t,~z_t)$$

Note that here, we use $[[\cdot]]$ to denote a concatenation along the time dimension. Otherwise, we will use $[\cdot]$ later to denote a concatenation along the features dimension. \\

\begin{minipage}{\linewidth}
\section{Model Architecture}

\subsection{Multi-Head Attention Mechanism}

For a concrete implementation of the previous formula, a modified version of the Multi-Head Attention can be used, similarly as seen in Attention Is All You Need \cite{vaswani2017attention}, with optionally some positional embedding to relatively index the keys back in time. Note that the positional embedding here performs a concatenation on the features rather than an addition as originally, a bit like a dense layer \cite{huang2017densely} rather than a residual  layer\cite{he2016deep}: 

$$v_t = [[c_{t−1},~c_{t−2},~c_{t−3},~...,~c_{t−k}]]$$
$$key=value=positionalEncoding(v_t)$$
$$query=W_{xh}([x_t,~h_{t-1}])$$
$$BNELU_j(arg)=BatchNorm_j(elu(arg))=BN(elu(arg))$$
$$a_t=MultiHeadSoftmax\bigg(\frac{query*BNELU_1(key)}{sqrt(d_k)}\bigg)*BNELU_2(values)$$
$$c_t = C(h_t,~a_t,~z_t)$$

Where $d_k$ is the dimensionality of every attention head: $$d_k=numberFeatures/numberHeads=2$$ The Batch Normalization is always 1-dimensional and applied on the features dimension, which requires a temporary flatten from a rank 3 tensor to a rank 2 tensor.

\end{minipage}

\begin{figure}
\centering
\includegraphics[width=1\textwidth]{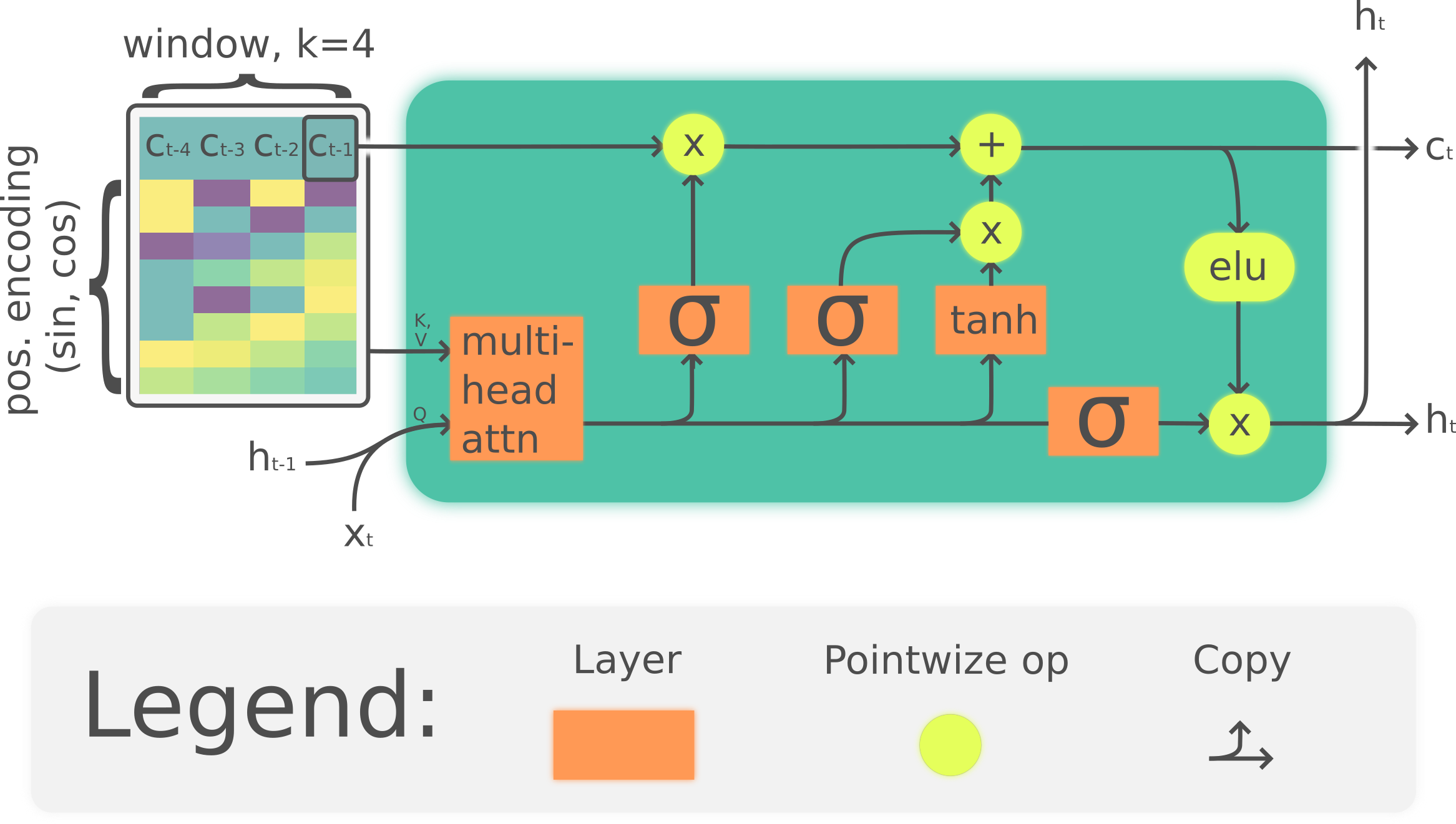}
\caption{\label{fig:larnn}A concrete implementation of the LARNN and its usage of the multi-head attention mechanism on a windowed queue of its past cell states. K, V and Q are respectively the attention mechanism's input Keys, Values and only one Query. The most recent states $c_{[t-1..t-k]}$ are arranged in a "first in, first out" queue of length $k$. There is also the not-illustrated fact that the query is formulated from a layer of the concatenation of $h_{t-1}$ and $x_t$ on the innermost (feature) axis. This visualization is freely available and is licensed under the CC-BY License, by Guillaume Chevalier. For more information, visit \url{https://github.com/guillaume-chevalier/Linear-Attention-Recurrent-Neural-Network/tree/master/inkscape_drawings}.}
\end{figure}

\begin{figure}
\centering
\includegraphics[width=0.9\textwidth]{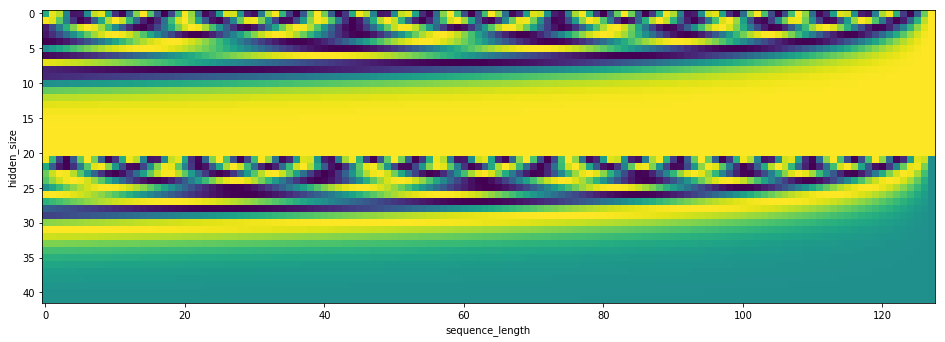}
\caption{\label{fig:original-attn}The original positional encoding used in Attention Is All You Need \cite{vaswani2017attention}, composed of sines and cosines. Note that here, the sines and cosine has been split rather than interpolated, and are used with a ${d_model}$ of 42 and a sequence length of 128.}
\end{figure}

\begin{figure}
\centering
\includegraphics[width=0.9\textwidth]{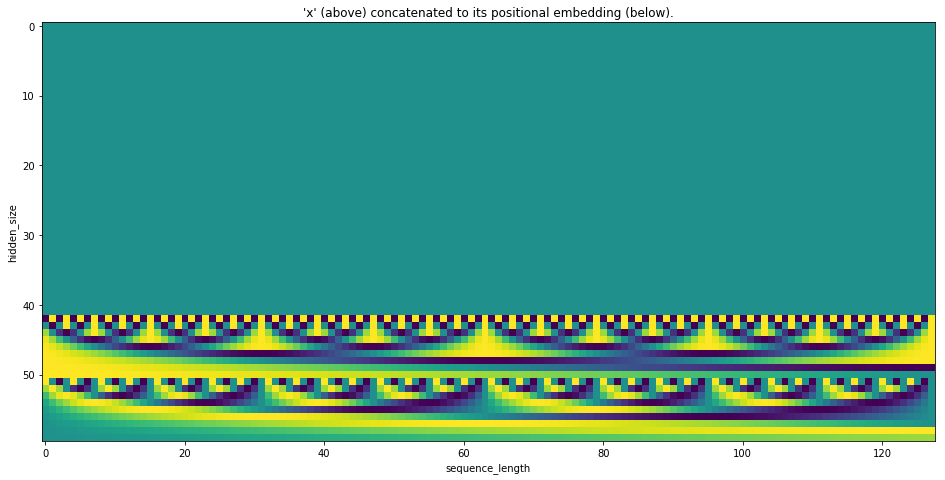}
\caption{\label{fig:concat-my-attn}The windowed cell states $c_t$ are here represented above, concatenated on the sequence length (window size) temporal axis, whereas the encoding is concatenated below on the feature size axis. That is, there are 42 features in the cells' states $c_t$ before the concatenation of the encoding which augments the number of such features to be processed by the multi-head attention as a keys and values. This visualization and the three previous ones are computed from the code of the project which can be found here: \url{https://github.com/guillaume-chevalier/Linear-Attention-Recurrent-Neural-Network/blob/master/AnnotatedMultiHeadAttention.ipynb}}
\end{figure}

\begin{minipage}{\linewidth}

\subsection{Special usage of Positional Encoding}

The positional encoding discussed in Attention Is All You Need \cite{vaswani2017attention} uses a geometric series of sines and cosines defined by the following formula and as represented in Figure \ref{fig:original-attn}:

$$PE_{(pos,~2i)} = sin(pos / 10000^{2i/d_{model}})$$
$$PE_{(pos,~2i+1)} = cos(pos / 10000^{2i/d_{model}})$$

where $d_{model}$ is here the same as the hidden unit size, that is, the number of features in the LARNN, akin the number of features (hidden size) in the LSTM. Their positional embedding may contain a random phase so as to let their Neural Machine Translation (NMT) model generalize to unknown sentences length. \\

In the case of the LARNN, no random phase is needed, because the LARNN always have a window of a fixed size. Therefore, the positional encoding is reversed and is therefore applied from the most recent cell state as being the zero of the sines and cosines, then positively towards older cell states $c_t$ (more details in the LSTM and LARNN equations later). Also, because the window size is fixed, it is no longer needed to express the encoding in function of the number of features, but now rather in function of the window size. In the case of the LARNN, perfect exponents of two were used for the wavelenghts, which yields its particularly "pixel-perfect" agreeable aspect, rather than using multiples of 1000, which was originally used in Attention Is All You Need for a reason that seems unknown reason to the best of my knowledge. \\

In Figure \ref{fig:concat-my-attn}, it's also possible to see that in the LARNN, the positional encoding is concatenated to the features, rather than added to them as in Attention Is All You Need. This is to replicate the effect of a dense layer \cite{huang2017densely} rather than a residual  layer\cite{he2016deep}. \\
\end{minipage}

\begin{figure}
\centering
\includegraphics[width=0.9\textwidth]{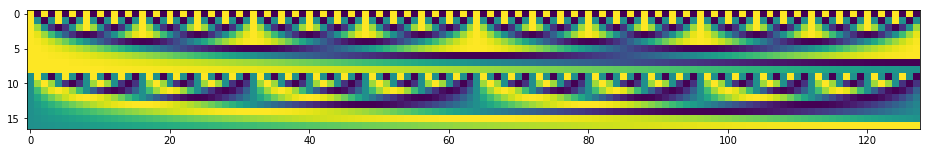}
\caption{\label{fig:my-attn}The modified positional encoding used in the LARNN, composed of sines and cosines. The window size is fixed to 128 for visualization purposes here, which causes the greatest wavelength to be of a quarter of its wavelength from zero to its first peak in case of a sine. Therefore, this type of encoding is such that its greatest wavelength is of at least four times the window size. In practice, for the LARNN, the window size is for sure smaller than $k=128$ to be linear, since in the used dataset the sequence length is of 128, which represents 2.56 seconds worth of data. From left to right would be placed, respectively, the most recent and the oldest windowed cell states $c_t$. }
\end{figure}

\begin{minipage}{\linewidth}

\subsection{Relationships with the Batch Normalized LSTM}

We now want to place the formulas of the LSTM RNN in place of the consciousness RNN $C$, such as in Figure \ref{fig:lstm}. For convenience, here are some formulas for an LSTM cell. Note that here, with a slight abuse of notation, $h_t$ is the output, and that $x_t$ is the input, whereas in the equations above, $h_t$ is instead the input.  Let's continue by using the LSTM equations with Batch Normalization (such as a BN-LSTM) \cite{cooijmans2016recurrent}, but here with an Exponential Linear Unit (ELU) activation \cite{clevert2015fast}, such as being:

$$f_t= ~~~~\sigma(BN(W_{hf}~h_{t-1} + W_{xf}~x_t + b_f))$$
$$i_t= ~~~~\sigma(BN(W_{hi}~h_{t-1}~+ W_{xi}~x_t+b_i))$$
$$o_t= ~~~~\sigma(BN(W_{ho}~h_{t-1} + W_{xo}~x_t+b_o))$$
$$g_t=       tanh(BN(W_{hc}~h_{t-1} + W_{xc}~x_t+b_c))$$
$$C_t=BN(f*C_{t-1} + i*{x_t}*g_t)$$
$$h_t=BN(o_t*elu(C_t))$$

\end{minipage}

\begin{minipage}{\linewidth}

\subsection{Putting it all together}

Let's now define how it's possible to merge the result of an attention query $a_t$ inside the LSTM formulas before incorporating this into $C$ function to replace it with the LSTM. In fact, two different ways can be defined as such, either one or the other - the residual or the layer mode of joining the attention into the LSTM cell - by replacing the definition of $g_t$ as either: 

$$g_{t~residual~mode} = 
tanh(BN(W_{hc}~h_{t-1} + W_{xc}~x_t + W_{ac}~a_t + b_c))$$
or:
$$g_{t~layer~mode} = Wa([x_t, h_{t-1}, a_t]) + b_a$$

With respect to the equations above to be merged and with the notation used in most of the LSTM formulas with $h_t$ to be the output and $x_t$ the input, here, $C$ is finally replaced by a BN-LSTM cell with attention to create the LARNN. It is possible to finally obtain the following equations, and with choosing optionally $g_{t~residual~mode}$ or $g_{t~layer~mode}$, which are both listed. We obtain something akin to what's seen in the Figure \ref{fig:larnn} which is an oversimplified representation of the LARNN implementation which goes like: 

$$v_t = [[c_{t−1},~c_{t−2},~c_{t−3},~...,~c_{t−k}]]$$
$$key=value=positionalEncoding(v_t)$$
$$query=W_{xh}([x_t,~h_{t-1}])$$
$$BNELU_j(arg)=BatchNorm_j(elu(arg))=BN(elu(arg))$$
$$a_t=MultiHeadSoftmax\bigg(\frac{query*BNELU_1(key)}{sqrt(d_k)}\bigg)*BNELU_2(values)$$
$$f_t= ~~~~\sigma(BN(W_{hf}~h_{t-1} + W_{xf}~x_t + b_f))$$
$$i_t= ~~~~\sigma(BN(W_{hi}~h_{t-1}~+ W_{xi}~x_t+b_i))$$
$$o_t= ~~~~\sigma(BN(W_{ho}~h_{t-1} + W_{xo}~x_t+b_o))$$
$$g_{t~residual~mode} = 
tanh(BN(W_{hc}~h_{t-1} + W_{xc}~x_t + W_{ac}~a_t + b_c))$$
$$g_{t~layer~mode} = Wa([x_t, h_{t-1}, a_t]) + b_a$$
$$C_t=BN(f*C_{t-1} + i*{x_t}*g_t)$$
$$h_t=BN(o_t*elu(C_t))$$

Note again that in the experiments here made, the BN-LSTM was modified such as to have an Exponential Linear Unit (ELU) activation \cite{clevert2015fast} in plus of Batch Normalization. This was a cheap inefficient trick to blindly try to obtain better results from an engineer's point of view. Note that such BN-ELU normalization was also added to the Multi-Head's linear mappings of the keys and values, which improved results too, as discussed in the analysis below.

\end{minipage}

\begin{figure}
\centering
\includegraphics[width=1\textwidth]{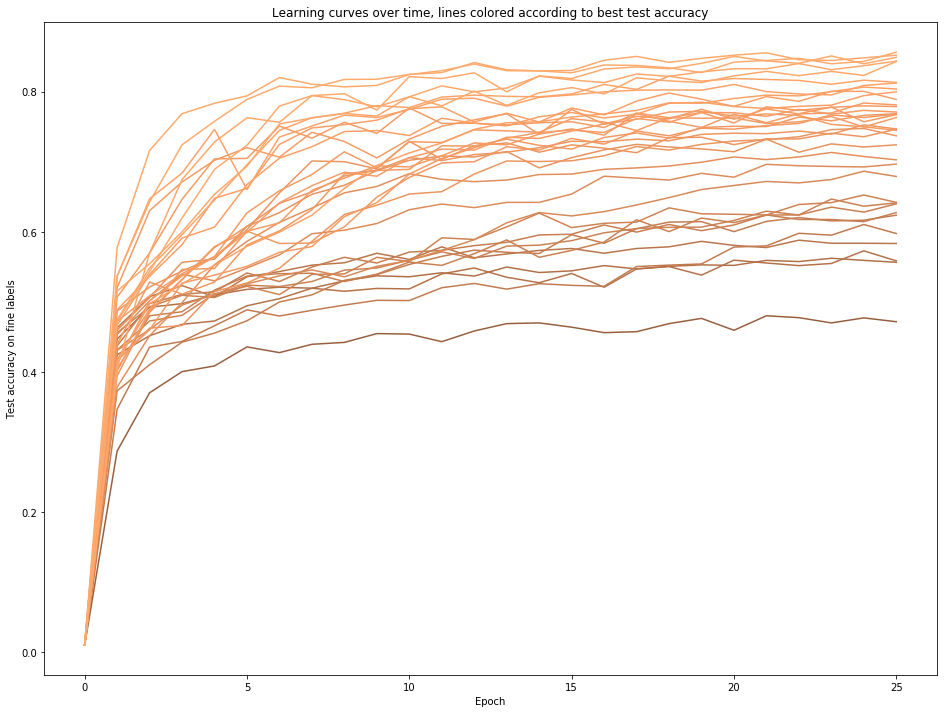}
\caption{\label{fig:r1}Plot illustrating the test accuracy of every trials through 25 epochs for the round 1 of hyperparameters optimization. For more information, visit \url{https://github.com/guillaume-chevalier/Linear-Attention-Recurrent-Neural-Network/blob/master/AnalyzeTestHyperoptResults_round_1.ipynb}.}
\end{figure}

\begin{figure}
\centering
\includegraphics[width=1\textwidth]{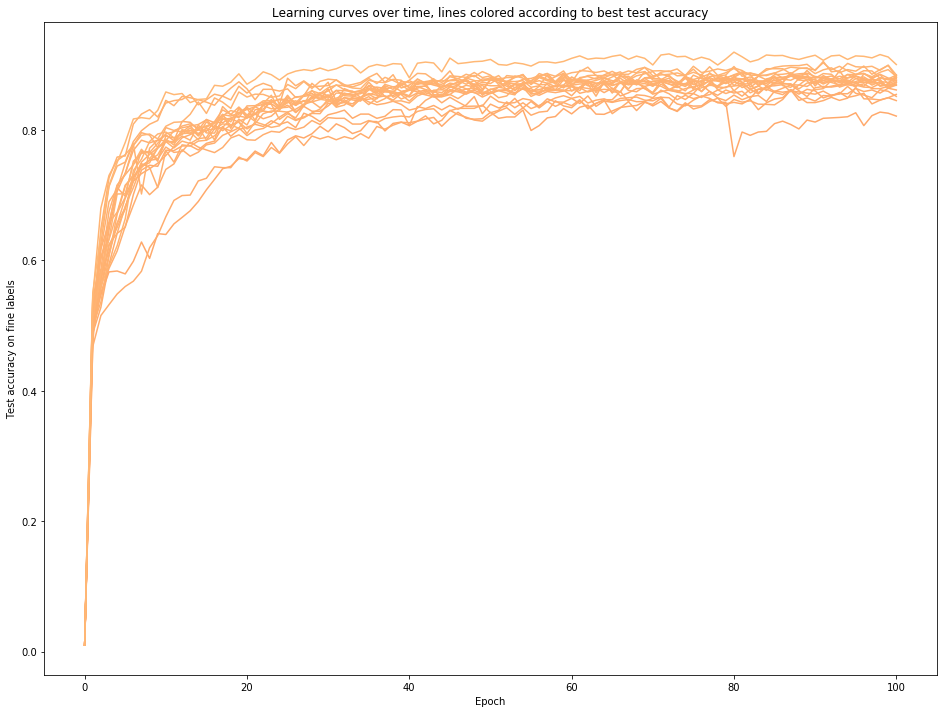}
\caption{\label{fig:r2}Plot illustrating the test accuracy of every trials through 100 epochs for the round 2 of hyperparameters optimization. For more information, visit \url{https://github.com/guillaume-chevalier/Linear-Attention-Recurrent-Neural-Network/blob/master/AnalyzeTestHyperoptResults_round_2.ipynb}.}
\end{figure}

\begin{figure}
\centering
\includegraphics[width=1\textwidth]{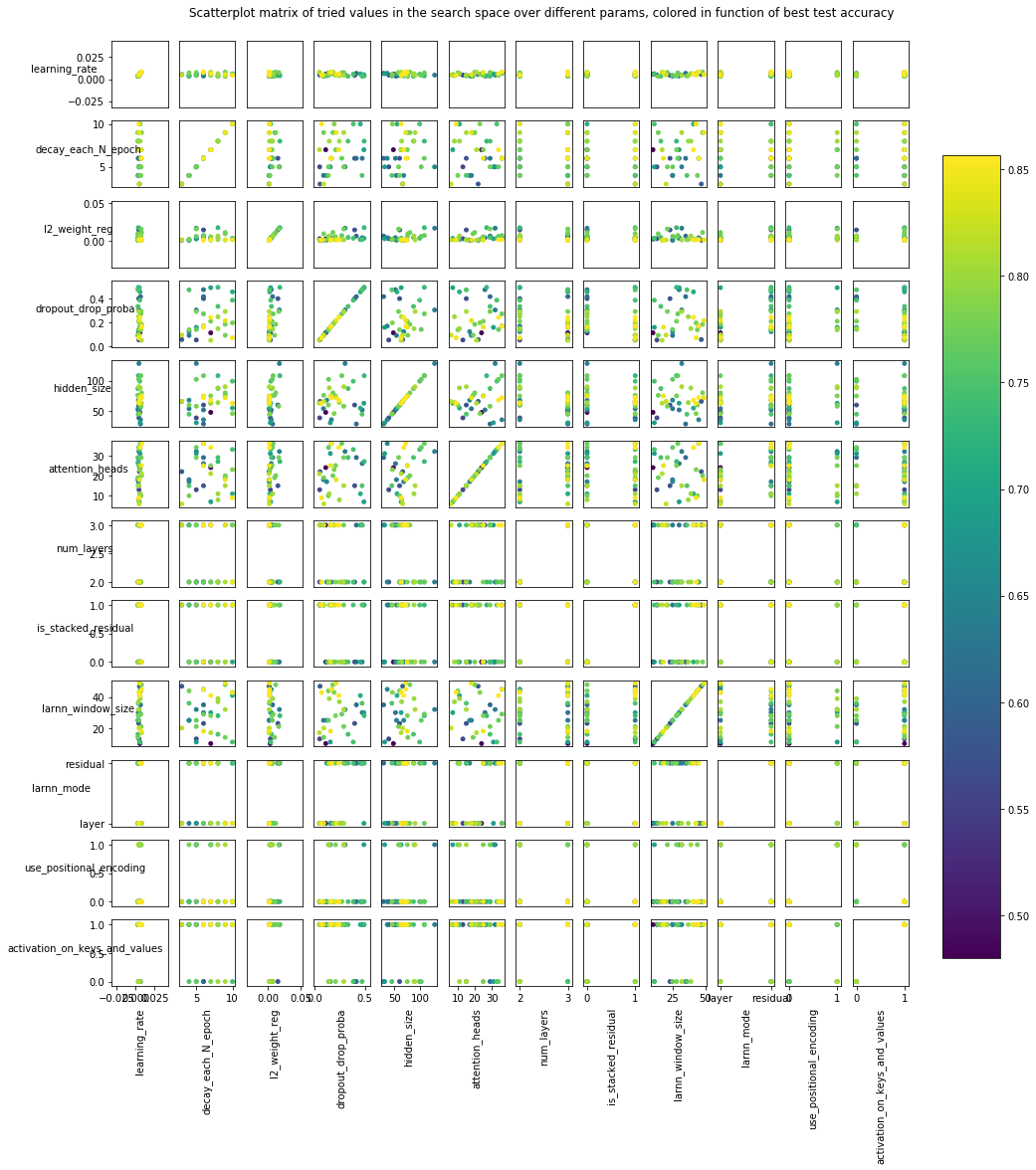}
\caption{\label{fig:r1h}Scatter plot depicting the effect of every hyperparameters in relation to each other for the optimization round 1. For more information, visit \url{https://github.com/guillaume-chevalier/Linear-Attention-Recurrent-Neural-Network/blob/master/AnalyzeTestHyperoptResults_round_1.ipynb}.}
\end{figure}

\begin{figure}
\centering
\includegraphics[width=1\textwidth]{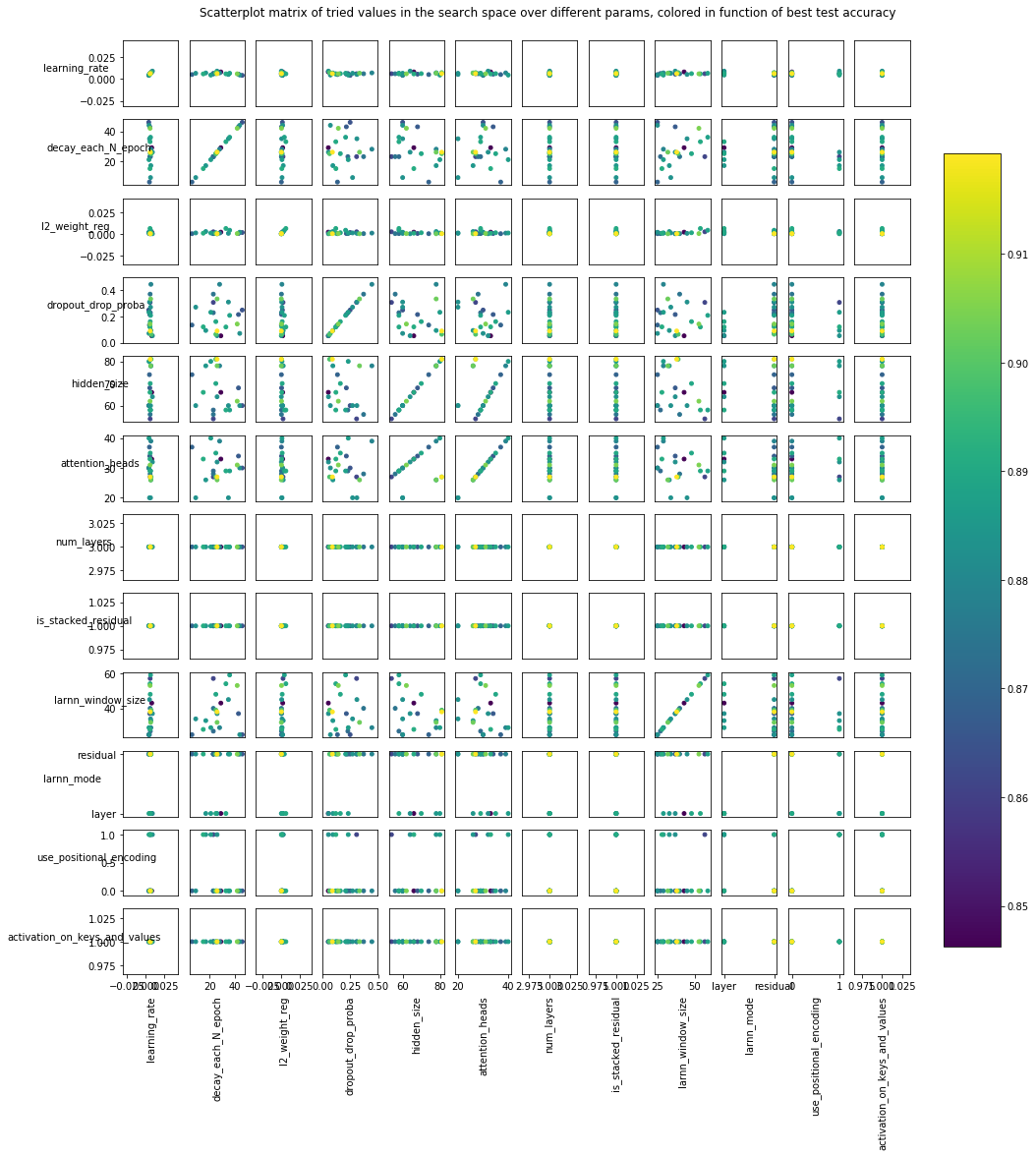}
\caption{\label{fig:r2h}Scatter plot depicting the effect of every hyperparameters in relation to each other for the optimization round 2. For more information, visit \url{https://github.com/guillaume-chevalier/Linear-Attention-Recurrent-Neural-Network/blob/master/AnalyzeTestHyperoptResults_round_2.ipynb}.}
\end{figure}

\begin{minipage}{\linewidth}

\section{Training Procedure and Results}

The training of the concrete implementation of the LARNN was performed with Hyperopt using the TPE algorithm \cite{bergstra2013hyperopt}, which is yields better results than a random hyperparameter search and a grid search \cite{bergstra2012random}. Two rounds of the TPE algorithm were performed. In the first round, many hyperparameters were set to vary a lot in order for the search to be in diffuse mode, and then later on in a second round, the search was set to a focused mode \cite{raichle2007default}, that is, by restraining the area of where the search is performed by fixing some hyperparameters to their best value or towards a good range of values. The dataset used was the Public Domain Dataset for Human Activity Recognition using Smartphones \cite{anguita2013public} by Anguita, Davide, et al., as uploaded on the UCI Machine Learning Repository. In this paper's code, this dataset is named the UCIHAR dataset. 

\subsection{Accuracies of Every Trained Model and Hyperparameters Analysis}

It is possible to see in Figure \ref{fig:r1} and in Figure \ref{fig:r2} the accuracies obtained at, respectively,  round 1 and round 2. Also, the effect of every hyperparameters related to each other are visualized in two scatter plots for the round 1 and the round 2 in Figure \ref{fig:r1h} and in Figure \ref{fig:r2h}. Those plots are not an ablation study, but are close to be. For example, sometimes different types of LARNN modes are used to create the layers, and sometimes the densely concatenated positional encoding is turned off. \\

The training was carried on Amazon AWS's p3.2xlarge instances, that is, with NVIDIA Tesla V100 GPUs of 16 GB RAM with 640 NVIDIA Tensor cores and 5'120 NVIDIA CUDA cores offering more than 100 TFLOPS. The total costs were of \$344.17 USD at a cost of \$3.06 USD per hour, which means 112 hours of GPU usage. \\
\end{minipage}

\begin{figure}
\centering
\includegraphics[width=1\textwidth]{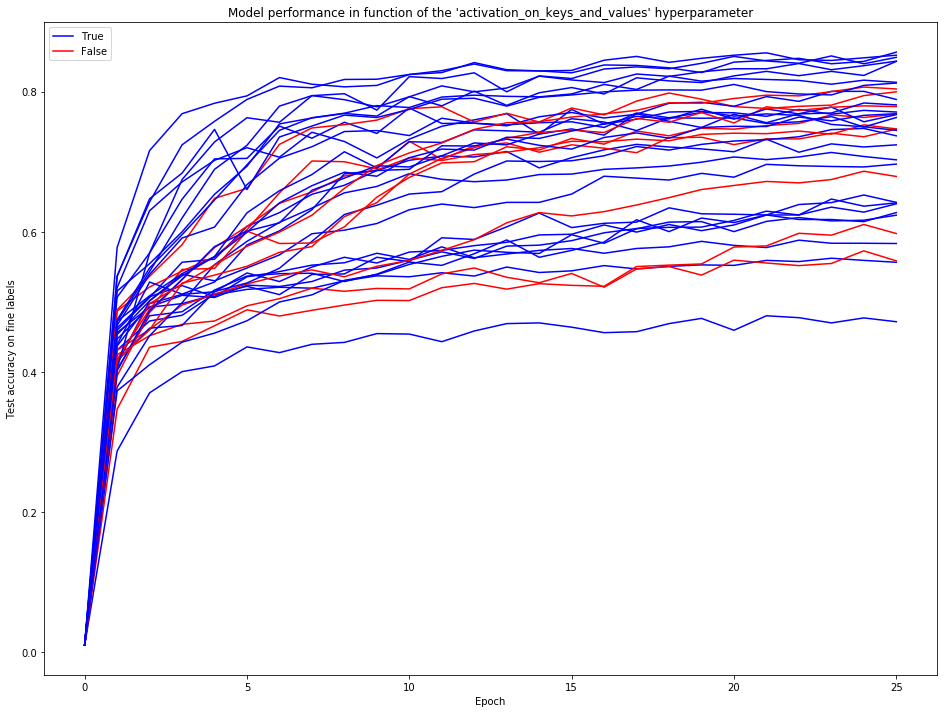}
\caption{\label{fig:activation}Test accuracy throughout training for when batch-normalized ELU activation is used on the keys and values' linear layers of the Multi-Head Attention Mechanism. This is an interesting discovery considering Multi-Head Attention Mechanisms are now, at the time of writing, the State Of The Art (SOTA) methods in solving Natural Language Processing (NLP) problems such as Neural Machine Translation (NMT), as done in 2017 in the Attention Is All You Need paper \cite{vaswani2017attention}, and bringing NLP systems closer to Artificial General Intelligence (AGI) such as in the One Model To Learn Them All paper \cite{kaiser2017one} using a Multi-Head Dot-Product Attention. However, note that the LARNN have still not been tested on NLP tasks as of today, it has only been tested on the sensors dataset used here. For more information on how this figure was generated, visit \url{https://github.com/guillaume-chevalier/Linear-Attention-Recurrent-Neural-Network/blob/master/AnalyzeTestHyperoptResults_round_1.ipynb}.}
\end{figure}

\begin{minipage}{\linewidth}

From the charts, it's possible to see that the positional encoding did not help. 
Also, it's better to stack the two or three cells in a residual fashion, such as to add their $h_t$ together at layer 1 and layer 2 before the final classification layer which is placed at the last time step. Also, it appears that the best LARNN mode is the "residual" one, while the "layer" one yields average results in the round 2. It's also possible to observe that the results are better when placing an activation on the Multi-Head Attention's linear mappings before the dot products and attention products are made, as seen in Figure \ref{fig:activation}. which is quite interesting and may orient future research when using Multi-Head Attention. As a reminder: the activation used is a Batch Normalization on an ELU activation of the linears. The Batch Normalization is 1-dimensional on the features dimension.\\

It can obtain results of 91.924\% for the test accuracy, compared to the previously attained 91.653\% using vanilla LSTM cells. \\

\end{minipage}

\begin{minipage}{\linewidth}

\section{Conclusion}

It is possible to infer new equations from the consciousness RNN and the Multi-Head Attention Mechanisms to create a LARNN which makes use the attention to augment the range of its consciousness. Moreover, intesresting hyperparameter exploration yields insights on what is useful to this neural architecture and what's not. For example, positional encoding did not seem to help specifically for this task if it is added in a dense manner, and also I discovered that batch-normalized activation on the linear mappings of the multi-head attention mechanism helped the model on the Test set predictions. \\

\section{Acknowledgements}

Thanks to Yu Zhao for having participated in the HAR-stacked-residual-bidir-LSTMs \cite{DBLP:journals/corr/abs-1708-08989} project with me, which is licensed under the Apache 2.0 license, a project from which I reused some code for loading the dataset. In turns, this code was originally derived from my own project available at \url{https://github.com/guillaume-chevalier/LSTM-Human-Activity-Recognition}.\\

Also, thanks to Vooban for open-sourcing and sublicensing under the MIT License its derivative of my code \url{https://github.com/guillaume-chevalier/Hyperopt-Keras-CNN-CIFAR-100} now at the address \url{https://github.com/Vooban/Hyperopt-Keras-CNN-CIFAR-100}, which I was able to reuse here as boilerplate code to setup the hyperparameter search. \\

More details on the licenses can be obtained at the respective address of each project. \\

I'd like to also thanks Philippe Giguère, Professor at Université Laval, who created a nice class on deep learning, such as listed in my awesome resources: \url{https://github.com/guillaume-chevalier/Awesome-Deep-Learning-Resources}. I especially liked his rare visualization of the multi-head attention mechanism, which is available in his slides at \url{https://ulaval-damas.github.io/glo4030/}, and more precisely, at the page 28 of his slides here: \url{http://www2.ift.ulaval.ca/~pgiguere/cours/DeepLearning/09-Attention.pdf}.\\

The LARNN repository is available at \url{https://github.com/guillaume-chevalier/Linear-Attention-Recurrent-Neural-Network}, available under the MIT License, and coded with PyTorch \cite{paszke2017automatic}.

\end{minipage}

\begin{minipage}{\linewidth}
\bibliographystyle{alpha}
\bibliography{sample}

\newcommand{\etalchar}[1]{$^{#1}$}
\begin{thebibliography}{VDODZ{\etalchar{+}}16}

\bibitem[AGO{\etalchar{+}}13]{anguita2013public}
Davide Anguita, Alessandro Ghio, Luca Oneto, Xavier Parra, and Jorge~Luis
  Reyes-Ortiz.
\newblock A public domain dataset for human activity recognition using
  smartphones.
\newblock In {\em ESANN}, 2013.

\bibitem[BB12]{bergstra2012random}
James Bergstra and Yoshua Bengio.
\newblock Random search for hyper-parameter optimization.
\newblock {\em Journal of Machine Learning Research}, 13(Feb):281--305, 2012.

\bibitem[BCB14]{bahdanau2014neural}
Dzmitry Bahdanau, Kyunghyun Cho, and Yoshua Bengio.
\newblock Neural machine translation by jointly learning to align and
  translate.
\newblock {\em arXiv preprint arXiv:1409.0473}, 2014.

\bibitem[Ben17]{bengio2017consciousness}
Yoshua Bengio.
\newblock The consciousness prior.
\newblock {\em arXiv preprint arXiv:1709.08568}, 2017.

\bibitem[BYC13]{bergstra2013hyperopt}
James Bergstra, Dan Yamins, and David~D Cox.
\newblock Hyperopt: A python library for optimizing the hyperparameters of
  machine learning algorithms.
\newblock In {\em Proceedings of the 12th Python in Science Conference}, pages
  13--20. Citeseer, 2013.

\bibitem[CBL{\etalchar{+}}16]{cooijmans2016recurrent}
Tim Cooijmans, Nicolas Ballas, C{\'e}sar Laurent, {\c{C}}a{\u{g}}lar
  G{\"u}l{\c{c}}ehre, and Aaron Courville.
\newblock Recurrent batch normalization.
\newblock {\em arXiv preprint arXiv:1603.09025}, 2016.

\bibitem[CUH15]{clevert2015fast}
Djork-Arn{\'e} Clevert, Thomas Unterthiner, and Sepp Hochreiter.
\newblock Fast and accurate deep network learning by exponential linear units
  (elus).
\newblock {\em arXiv preprint arXiv:1511.07289}, 2015.

\bibitem[HLWvdM17]{huang2017densely}
Gao Huang, Zhuang Liu, Kilian~Q Weinberger, and Laurens van~der Maaten.
\newblock Densely connected convolutional networks.
\newblock In {\em Proceedings of the IEEE conference on computer vision and
  pattern recognition}, volume~1, page~3, 2017.

\bibitem[HS97]{hochreiter1997long}
Sepp Hochreiter and J{\"u}rgen Schmidhuber.
\newblock Long short-term memory.
\newblock {\em Neural computation}, 9(8):1735--1780, 1997.

\bibitem[HZRS16]{he2016deep}
Kaiming He, Xiangyu Zhang, Shaoqing Ren, and Jian Sun.
\newblock Deep residual learning for image recognition.
\newblock In {\em Proceedings of the IEEE conference on computer vision and
  pattern recognition}, pages 770--778, 2016.

\bibitem[KGS{\etalchar{+}}17]{kaiser2017one}
Lukasz Kaiser, Aidan~N Gomez, Noam Shazeer, Ashish Vaswani, Niki Parmar, Llion
  Jones, and Jakob Uszkoreit.
\newblock One model to learn them all.
\newblock {\em arXiv preprint arXiv:1706.05137}, 2017.

\bibitem[PGC{\etalchar{+}}17]{paszke2017automatic}
Adam Paszke, Sam Gross, Soumith Chintala, Gregory Chanan, Edward Yang, Zachary
  DeVito, Zeming Lin, Alban Desmaison, Luca Antiga, and Adam Lerer.
\newblock Automatic differentiation in pytorch.
\newblock In {\em NIPS-W}, 2017.

\bibitem[RS07]{raichle2007default}
Marcus~E Raichle and Abraham~Z Snyder.
\newblock A default mode of brain function: a brief history of an evolving
  idea.
\newblock {\em Neuroimage}, 37(4):1083--1090, 2007.

\bibitem[VDODZ{\etalchar{+}}16]{van2016wavenet}
Aaron Van Den~Oord, Sander Dieleman, Heiga Zen, Karen Simonyan, Oriol Vinyals,
  Alex Graves, Nal Kalchbrenner, Andrew Senior, and Koray Kavukcuoglu.
\newblock Wavenet: A generative model for raw audio.
\newblock {\em arXiv preprint arXiv:1609.03499}, 2016.

\bibitem[VSP{\etalchar{+}}17]{vaswani2017attention}
Ashish Vaswani, Noam Shazeer, Niki Parmar, Jakob Uszkoreit, Llion Jones,
  Aidan~N Gomez, {\L}ukasz Kaiser, and Illia Polosukhin.
\newblock Attention is all you need.
\newblock In {\em Advances in Neural Information Processing Systems}, pages
  6000--6010, 2017.

\bibitem[ZYCG17]{DBLP:journals/corr/abs-1708-08989}
Yu~Zhao, Rennong Yang, Guillaume Chevalier, and Maoguo Gong.
\newblock Deep residual bidir-lstm for human activity recognition using
  wearable sensors.
\newblock {\em CoRR}, abs/1708.08989, 2017.

\end{thebibliography}
\end{minipage}


\end{document}